\documentclass{article}

\usepackage{arxiv}

\usepackage[utf8]{inputenc} 
\usepackage[T1]{fontenc}    
\usepackage{hyperref}       
\usepackage{url}            
\usepackage{booktabs}       
\usepackage{amsfonts}       
\usepackage{nicefrac}       
\usepackage{microtype}      
\usepackage{lipsum}		
\usepackage{graphicx}
\usepackage{natbib}
\usepackage{doi}
\usepackage{tabularx}
\usepackage{amsmath}

\title{Should We Trust This Summary? \\ Bayesian Abstractive Summarization to The Rescue}

\author{ \href{https://orcid.org/0000-0001-7639-6265}{\includegraphics[scale=0.06]{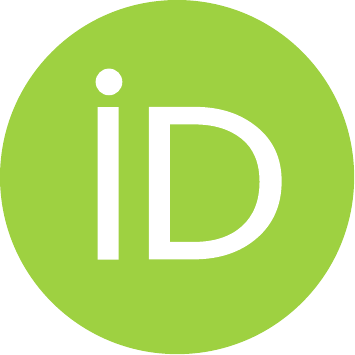}\hspace{1mm}Alexios Gidiotis} \\
	School  of  Informatics\\
	Aristotle University of Thessaloniki\\
	Thessaloniki, Greece 54124 \\
	\texttt{gidiotis@csd.auth.gr} \\
	\And
	\href{https://orcid.org/0000-0002-7879-669X}{\includegraphics[scale=0.06]{orcid.pdf}\hspace{1mm}Grigorios Tsoumakas} \\
	School  of  Informatics\\
	Aristotle University of Thessaloniki\\
	Thessaloniki, Greece 54124 \\
	\texttt{greg@csd.auth.gr} \\
}

\renewcommand{\headeright}{Technical Report}
\renewcommand{\undertitle}{Technical Report}
\renewcommand{\shorttitle}{Uncertainty-Aware Abstractive Summarization}

\hypersetup{
pdftitle={A template for the arxiv style},
pdfsubject={cs.cl},
pdfauthor={Alexios Gidiotis, Grigorios Tsoumakas},
pdfkeywords={abstractive summarization, Bayesian inference, uncertainty, Monte Carlo dropout},
}

\begin{document}
\maketitle

\begin{abstract}
	We explore the notion of uncertainty in the context of modern abstractive summarization models, using the tools of Bayesian Deep Learning. Our approach approximates Bayesian inference by first extending state-of-the-art summarization models with Monte Carlo dropout and then using them to perform multiple stochastic forward passes. Based on Bayesian inference we are able to effectively quantify uncertainty at prediction time. Having a reliable uncertainty measure, we can improve the experience of the end user by filtering out generated summaries of high uncertainty. Furthermore, uncertainty estimation could be used as a criterion for selecting samples for annotation, and can be paired nicely with active learning and human-in-the-loop approaches. Finally, Bayesian inference enables us to find a Bayesian summary which performs better than a deterministic one and is more robust to uncertainty. In practice, we show that our Variational Bayesian equivalents of BART and PEGASUS can outperform their deterministic counterparts on multiple benchmark datasets.
\end{abstract}


\section{Introduction}
\label{sec:introdction}
State-of-the-art text summarization methods have achieved remarkable performance in various benchmarks \cite{Song2019MASS:Generation, Dong2019UnifiedGeneration, Lewis2019BART:Comprehension, Zhang2020PEGASUS:Summarization}. The majority of these methods use very large Transformer models pre-trained on language generation tasks.

Although such methods can generate high quality summaries for texts similar to their training set, they suffer from a couple of issues when the inputs lie far from the training data distribution. They are prone to generating particularly bad outputs \cite{Xu2020UnderstandingUncertainty,Kryscinski2020EvaluatingSummarization} and are usually fairly confident about them \cite{Gal2016DropoutLearning, Xiao2020WatTransformers}. These shortcomings are bound to cause problems once a summarization model is deployed to solve a practical problem.

Since the output of automatic summarization models is usually expected to be consumed by humans, it is very important to know when such an output is of good enough quality to be served to users. In most cases, it is very much preferable to not serve an output at all, instead of serving a bad output. This will in turn increase users’ trust to automated summarization systems.

Model uncertainty is one way of detecting when a model's output is likely to be poor on the grounds of predicting far away from it’s training distribution. Recent summarization methods have focused heavily on improving the overall performance, but model uncertainty has been explored very little \cite{Xu2020UnderstandingUncertainty}. 

In addition to improving user experience, the development of uncertainty measures for summarization can pave the way for active learning approaches \cite{Gal2017DeepData, Houlsby2011BayesianLearning, Liu2020LTP:Recognition, Lyu2020YouAnswering}. The value of active learning stems from the fact that obtaining labeled samples for training is hard, but it is relatively easy to obtain large amounts of unlabeled samples. Summarization is no different in this perspective, since creating good quality target summaries for training can be very costly.

This work explores uncertainty estimation for state-of-the-art text summarization models, from a Bayesian perspective. We extend the BART \cite{Lewis2019BART:Comprehension} and PEGASUS \cite{Zhang2020PEGASUS:Summarization} summarization models with Monte Carlo dropout \cite{Gal2016DropoutLearning}, in order to create corresponding Variational Bayesian PEGASUS (VarPEGASUS) and BART (VarBART) models. Sampling multiple summaries from those models allows us to approximate Bayesian inference in a practical way, which in turn enables us to estimate summarization uncertainty. To the best of our knowledge this is the first attempt to apply Bayesian summary generation with large Transformer models.

Based on Bayesian approximation, we adapt the Monte Carlo BLEU variance metric \cite{Xiao2020WatTransformers} to the summarization task, and investigate its efficacy as a measure of summarization uncertainty. Our findings suggest that this uncertainty metric correlates well with the quality of the generated summaries and can be effective at identifying cases of questionable quality.

Finally, we take the summarization uncertainty study one step further, and select the summary with the lowest disagreement out of multiple summaries sampled from our Variational models. Experiments across multiple benchmark datasets show that this method consistently improves summarization performance (see Table \ref{tab:good_sample}), and by using it our VarPEGASUS and VarBART models achieve better ROUGE F-scores compared to their original deterministic counterparts.

The rest of this paper is structured as follows. Section \ref{sec:related_work} discusses related work on Bayesian deep learning and uncertainty estimation methods. Section \ref{sec:methods} presents our approach. Section \ref{sec:experiments} describes our experimental setup, while Section \ref{sec:results} presents and discusses the results. Finally, Section \ref{sec:conclusion} concludes our work and considers its broader impact.

\section{Related work}
\label{sec:related_work}
Uncertainty estimation in deep learning is a topic that has been studied extensively. Bayesian deep learning includes a family of methods that attempt to capture the notion of uncertainty in deep neural networks. Such methods have gained increased popularity in the deep learning literature and there exist multiple applications in subfields such as Computer Vision \cite{Kendall2017WhatVision, Litjens2017AAnalysis, Gal2017DeepData} and Natural Language Processing (NLP)  \cite{Siddhant2020DeepStudy, Liu2020LTP:Recognition, Lyu2020YouAnswering, Xiao2020WatTransformers}.

Despite their obvious advantage of modeling uncertainty, the main problem with Bayesian deep learning methods is the computational cost of full Bayesian inference. To tackle this problem, \citet{Gal2016DropoutLearning} propose using standard dropout \cite{Srivastava2014Dropout:Overfitting} as a practical approximation of Bayesian inference in deep neural networks and call this method Monte Carlo dropout. \citet{Gal2017DeepData} use a convolutional neural network with Monte Carlo dropout in order to obtain an uncertainty estimate for active learning in the task of image classification. \citet{Houlsby2011BayesianLearning} sample many networks with Monte Carlo simulation and propose an objective function that takes into account the disagreement and confidence of the predictions coming from these networks.

Similar methods have also been applied to NLP. In machine translation, \citet{Xiao2020WatTransformers} extend the Transformer architecture with MC dropout to get a Variational Transformer, and use it to sample multiple translations from the approximate posterior distribution. They also introduce BLEUVar, an uncertainty metric based on the BLEU score \cite{Papineni2002BLEU:Translation} between pairs of the generated translations. \citet{Lyu2020YouAnswering} extend the work of \citet{Xiao2020WatTransformers} to question answering and propose an active learning approach based on a modified BLEUVar version. Similarly, \citet{Liu2020LTP:Recognition} use a conditional random field to obtain uncertainty estimates for active learning and apply their method to named entity recognition.

Although summarization is a prominent NLP task, summarization uncertainty has not been widely studied. \citet{Xu2020UnderstandingUncertainty} is the only work that focuses on uncertainty for summarization, but their work does not make use of Bayesian methods. They define a generated summary’s uncertainty based on the entropy of each token generated by the model during the decoding phase. Their study includes experiments on CNN/DM and XSum using the PEGASUS and BART summarization models. Their main focus is on understanding different properties of uncertainty during the decoding phase, and their work is not directly comparable to ours.

\section{Methods}
\label{sec:methods}
We first introduce Bayesian inference, in the context of deep neural networks and show how it can be used to measure uncertainty. Subsequently, we show how Bayesian inference can be applied to summarization in order to estimate the uncertainty of a summary generated for a given input. Finally, we show how Bayesian inference can be employed for producing better summaries.

\subsection{Monte Carlo dropout}
Contrary to standard neural networks, Bayesian probabilistic models capture the uncertainty notion explicitly. The goal of such models is to derive the entire {\em posterior} distribution of model parameters $\theta$ given training data $X$ and $Y$ (Equation \ref{eq:posterior}).

\begin{equation}
\label{eq:posterior}
P(\theta | X,Y) = \frac{P(Y | X, \theta)P(\theta)}{P(Y | X)}
\end{equation}

At test time, given some input $x$, a prediction $\hat{y}$  can be made by integrating over all possible $\theta$ values (Equation \ref{eq:integration}). The predictive distribution’s variance can then be used as a measure of the model's uncertainty.

\begin{equation}
\label{eq:integration}
P(\hat{y} | x,X,Y) = \int P(\hat{y} | x, \theta) P(\theta | X,Y) d\theta
\end{equation}

In practice, integrating over all possible parameter values for a deep neural network is intractable, and therefore Variational methods are used to approximate Bayesian inference. A neural network trained with dropout can be interpreted as a Variational Bayesian neural network \cite{Gal2016DropoutLearning}, and as a result making stochastic forward passes with dropout turned on at test time is equivalent to drawing from the model's predictive distribution. This Monte Carlo (MC) dropout method can be easily applied to any neural network that has been trained with dropout.

\subsection{Summary uncertainty}
MC dropout is a simple yet effective method that requires no adjustment to the underlying model. It is possible to convert any state-of-the-art summarization model to a {\em Variational Bayesian model}, with the use of MC dropout. For Transformer based models in particular, the Transformer blocks that make up the encoder and decoder are usually trained with dropout, and therefore the conversion is trivial by simply turning dropout on at test time.

In Variational models, the variance of the predictive distribution can be used to measure the model’s uncertainty. For a text summarization model, we can approximate the variance of this distribution, by measuring the dissimilarity of $N$ stochastic summaries $y_1,y_2 \ldots y_N$, generated with MC dropout.

The BLEU metric \cite{Papineni2002BLEU:Translation} is commonly used for measuring the similarity between a pair of texts. As in \citet{Xiao2020WatTransformers}, we approximate the model’s predictive variance with the BLEU Variance (BLEUVar) metric over the $N$ summaries generated with MC dropout as shown in Equation \ref{eq:bleuvar}. BLEUVar is computed by summing the squared complement of BLEU among all pairs of summaries (twice as BLEU is asymmetric) generated for the same input with different dropout masks.

\begin{equation}
\small
\label{eq:bleuvar}
\textrm{BLEUVar} = \sum_{i=1}^{N} \sum_{j \neq i}^{N} (1 - \textrm{BLEU}(y_i,y_j))^2
\end{equation}

Because we sum over all pairs of $N$ samples twice, scores that are computed with different $N$ values are not directly comparable. To alleviate this issue we propose a normalized version of the metric, BLEUVarN, where we divide BLEUVar by $N(N-1)$ (Equation \ref{eq:norm_bleuvar}). This allows for comparisons between scores computed with different $N$ values.

\begin{equation}
\small
\label{eq:norm_bleuvar}
\textrm{BLEUVarN} = \frac{\sum_{i=1}^{N} \sum_{j \neq i}^{N} (1 - \textrm{BLEU}(y_i,y_j))^2}{N(N-1)}
\end{equation}

By running multiple stochastic forward passes for the same input, we essentially create an ensemble of models with different parameters. Making predictions with this ensemble has the following effects. For inputs close to the learned distribution the summaries generated by all models in the ensemble will be similar to one another, and as a result BLEUVarN will be low. On the other hand, for inputs lying away from the learned distribution, the generated summaries will differ wildly and BLEUVarN will be high, indicating high uncertainty.

\subsection{Bayesian summary generation}
\label{subsec:bayes_inference}
Inspired by the fact that making multiple predictions with MC dropout is equivalent to ensembling multiple stochastic models, we propose a novel Bayesian approach to summary generation. Instead of generating a single deterministic summary without dropout, as is commonly the case with modern summarization approaches, we consider using the {\em predictive mean} of multiple predictions made with MC dropout. Because the predictions in our case are summaries their predictive mean is not well defined, so instead we opt for selecting one of the $N$ summaries.

We assume that the {\em predictive mean} of the $N$ summaries generated with MC dropout should be the one having the lowest {\em disagreement} with the rest of the $N-1$ summaries. Since the pairwise complement of BLEU between all pairs of the sampled summaries has already been computed when estimating BLEUVarN uncertainty, it can be further used to help us find the lowest disagreement summary. In practice, we  select the summary $\hat{\mu}$ that maximizes the sum of symmetric BLEU similarity with the rest of the summaries (Equation \ref{eq:median}) \cite{Xiao2020WatTransformers}. This summary could be seen as the {\em median} of all the summaries generated with MC dropout, although this is not a mathematically correct expression. 

\begin{equation}
\small
\label{eq:median}
\hat{\mu} = \underset{y_i}{\mathrm{argmax}} \sum_{j \ne i}^{N}\left[\textrm{BLEU}(y_i,y_j) +\textrm{BLEU}(y_j,y_i)\right]
\end{equation}

The intuition behind this approach is based on the following assumption. We expect the {\em median} summary to integrate the key concepts that all individual summaries agree on. Consequently, for inputs close to the model's learned distribution, the individual summaries will be similar to one another and as a result the {\em median summary} will be the best choice. On the other hand, for out-of-distribution inputs, the {\em median} out of a number of very different summaries will result in a more robust and overall better final summary. In practice, even for well trained models, we expect to have a fairly large number of inputs that are not close to the models’ learned distribution, and therefore we expect to benefit from the positive effects of ensembling multiple outputs.

\section{Experimental Setup}
\label{sec:experiments}

We first present the three datasets that are involved in our experiments, their main statistics and the reasons for including them in our empirical study. Then we present the two summarization models that we employed, along with their parameters and details on stochastic summary generation. 

\subsection{Datasets}
In order to verify the effectiveness of our Bayesian abstractive summarization approach, we conducted a series of experiments on three well-known summarization benchmarks: 
\begin{itemize}
    \item {\bf XSum} \cite{Narayan2018DontSummarization} is a dataset of 227k BBC articles on a wide variety of topics. Each article is accompanied by a human written, single-sentence summary.
    \item {\bf CNN/DailyMail} \cite{Hermann2015TeachingComprehend} is a dataset containing a total of 93k articles from the CNN, and 220k articles from the Daily Mail newspapers. All articles are paired with bullet point summaries. The version used is the non-anonymized variant similar to \cite{See2017GetNetworks}. 
    \item {\bf AESLC} \cite{Zhang2020ThisGeneration} is a dataset of 18k emails from the Enron corpus \cite{Klimt2004TheResearch}. The body of each email is used as source text and the subject as summary.  
\end{itemize}

The main criteria for selecting these datasets are the availability of recent, open source models trained on them and their relatively short texts that would allow us to run a number of different experiments quickly. Since our methods do not involve training, we will only focus on the validation and test set of each dataset. All datasets are obtained from the Hugging Face datasets repository\footnote{\href{https://huggingface.co/datasets}{https://huggingface.co/datasets}}. Table \ref{tab:data_stats} presents some basic statistics for these datasets.

\begin{table}[t!]
\caption{\label{tab:data_stats} Basic statistics for each one the datasets used in our experiments. The document and summary length are measured in words.}
\begin{center}
\begin{tabular}{lrrrr}
\hline
 & \multicolumn{2}{c}{{\bf Size}} & \multicolumn{2}{c}{{\bf Length}} \\
{\bf Dataset} & {\bf Val.} & {\bf Test} & {\bf Doc.} & {\bf Sum.} \\
\hline
XSum & 11,332 & 11,334 & 431 & 23 \\
CNN/DM & 13,368 & 11,490 & 760 & 46 \\
AESLC & 1,960 & 1,906 & 75 & 4 \\
\hline
\end{tabular}
\end{center}
\end{table}

\subsection{Models}
BART \cite{Lewis2019BART:Comprehension} and PEGASUS \cite{Zhang2020PEGASUS:Summarization} are Transformer based sequence-to-sequence models, pre-trained on massive corpora of unsupervised data (Web and news articles). Since our experiments do not involve training, we utilize open-source models fine-tuned on the training set of each dataset. These models can be found in the Hugging Face models repository\footnote{\href{https://huggingface.co/models}{https://huggingface.co/models}}.

Our BART models follow the BART\textsubscript{LARGE} architecture with 12 Transformer blocks for the encoder and the decoder. BART is pre-trained as a denoising autoencoder, where the text is corrupted and the model learns to reconstruct the original text. Open-source fine-tuned BART models are only available for XSum and CNN/DM. Our PEGASUS models follow the PEGASUS\textsubscript{LARGE} architecture and have 16 Transformer blocks for the encoder and the decoder. PEGASUS is pre-trained on the C4 and HugeNews datasets, on a sentence infilling task. Open-source fine-tuned PEGASUS models exist for all three datasets considered in our experiments.


In order to convert BART and PEGASUS to Variational models, we enable dropout for all Transformer blocks of the encoder and decoder. For each sample, we generate $N$ summaries using beam search decoding with 8 beams. We experimented with $N$ equal to $10$ (MC-10) and $20$ (MC-20). The rest of the hyper-parameters used were identical to the original papers.

\section{Results}
\label{sec:results}
Our main experiment evaluates BLEUVarN's effectiveness in quantifying uncertainty for summarization models. A second experiment investigates the potential of the Bayesian summarization method proposed in Section \ref{subsec:bayes_inference} as a way of improving summarization performance at test time.

\subsection{Evaluating Bayesian uncertainty}\label{subsec:bleuvar_exp}
We here evaluate the effectiveness of BLEUVarN in measuring the model's uncertainty. The performance versus data retention curve \cite{Filos2019ATasks} measures how well a model would perform if we completely removed the $k$\% most uncertain outputs from the test set. In the $x$-axis we have the fraction of data from the test set that are removed, while in the $y$-axis we have the performance metrics. An effective uncertainty measure should show a consistent improvement in performance as we discard more samples based on high uncertainty. In this experiment, we arrange samples by decreasing BLEUVarN score and gradually remove the samples with the highest score.

\begin{figure*}[!t]
    \makebox[\textwidth][c]{\includegraphics[width=1.2\textwidth]{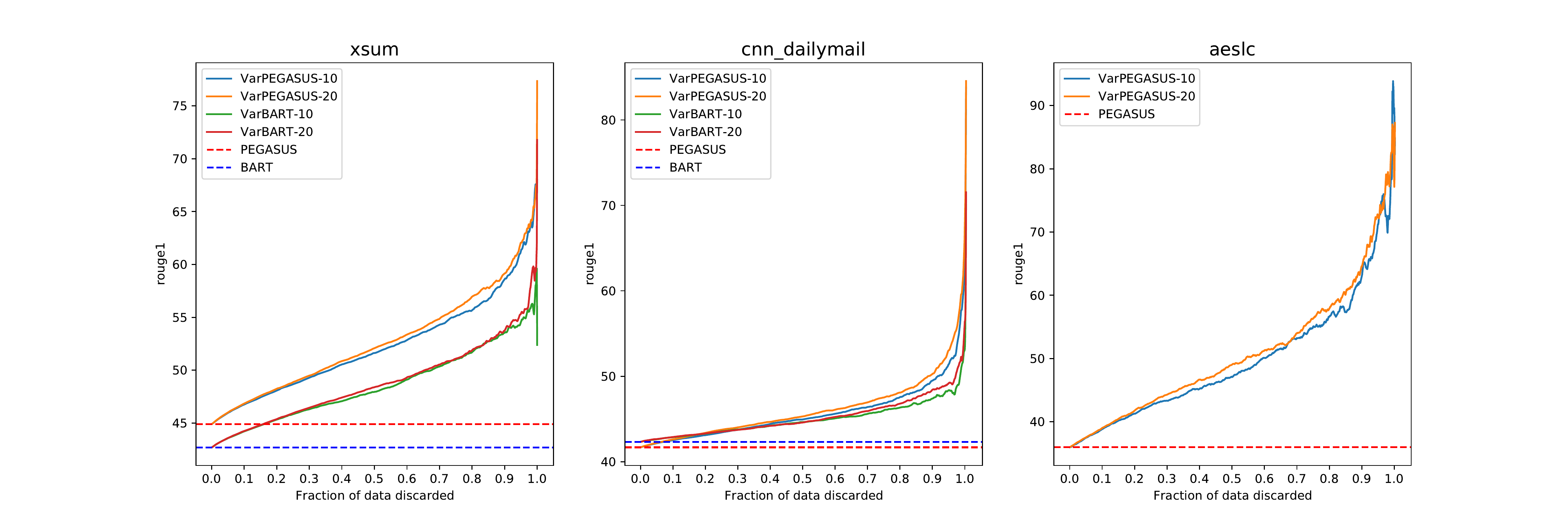}}
    \caption{ROUGE-1 scores vs fraction of data discarded due to high BLEUVarN. The straight dashed lines indicate the performance level of the deterministic PEGASUS and BART models.}
    \label{fig:r1_data}
\end{figure*}

Figure \ref{fig:r1_data} shows, for each dataset, the performance of our Variational models in terms of ROUGE-1 F-score versus the fraction of data discarded based on BLEUVarN. ROUGE-2 and ROUGE-L F-scores follow similar patterns and can be found in the Appendix \ref{sec:appendix}. For reference, we are also plotting the performance of the deterministic models using all data as straight lines. Also, in Table \ref{tab:percent_increase} we quantify the percentage increase in ROUGE F-scores as we discard different fractions of the full test datasets based on BLEUVarN.

\begin{table}[!t]
\caption{\label{tab:percent_increase} Percentage increase in ROUGE F-scores when discarding 25\%, 50\% and 75\% of the data based on the highest BLEUVarN.}
\begin{center}
\begin{tabular}{rccc}
\hline
& {\bf 25\%} & {\bf 50\%} & {\bf 75\%}\\
{\bf Model} & {\bf R-1/R-2/R-L} & {\bf R-1/R-2/R-L} & {\bf R-1/R-2/R-L}\\
\hline
& \multicolumn{3}{c}{\textbf{XSum}}\\
\hline
VarBart-10 & 6.4/13.8/8.3 & 12.2/25.2/15.4 & 22.1/41.9/26.1\\
VarBart-20 & 6.5/14.1/8.3 & 13.2/26.7/16.5 & 22.5/42.6/27.2\\
VarPegasus-10 & 7.5/15.8/9.6 & 14.9/29.4/18.6 & 25.2/4.9/29.9\\
VarPegasus-20 & 8.0/16.8/10.3 & 15.8/31.2/19.6 & 26.3/48.1/31.2\\
\hline
& \multicolumn{3}{c}{\textbf{CNN/DM}}\\
\hline
VarBart-10 & 2.9/7.2/4.8& 5.4/13.1/8.5& 8.8/20.4/13.3\\
VarBart-20 & 3.2/7.8/5.1& 5.3/12.8/8.5& 8.3/19.4/12.6\\
VarPegasus-10 & 4.1/9.9/6.1& 7.8/17.4/10.9& 12.6/26.1/16.8\\
VarPegasus-20 & 4.6/10.7/6.8& 8.5/19.0/11.9& 14.7/29.6/18.7\\
\hline
& \multicolumn{3}{c}{\textbf{AESLC}}\\
\hline
VarPegasus-10 & 17.5/33.5/17.7& 30.6/51.9/31.1& 54.4/75.0/54.7\\
VarPegasus-20 & 18.7/36.3/18.9& 36.0/59.7/36.6& 58.4/78.0/58.8\\
\hline
\end{tabular}
\end{center}
\end{table}

All ROUGE F-scores improve as we gradually discard samples with high BLEUVarN, an observation that is consistent across all test datasets and models. More specifically, we notice that the increase is linear for the first $80\%$ of the data, but then becomes almost exponential. From these observations we can draw two conclusions. First, models indeed perform significantly worse on samples with high uncertainty. Second, BLEUVarN is effective at quantifying uncertainty and can be used to identify high uncertainty samples.

Furthermore, we notice that the performance increase is significantly higher in the XSum and AESLC datasets compared to CNN/DM. In particular, VarPEGASUS-20 shows a staggering 58 point increase in ROUGE-1 score. We think that this difference might be related to the more extractive nature of CNN/DM summaries as opposed to the other two datasets. Such a finding would mean that Bayesian uncertainty filtering is more beneficial in abstractive rather than extractive setups.

\begin{figure*}[!t]
    \makebox[\textwidth][c]{\includegraphics[width=1.2\textwidth]{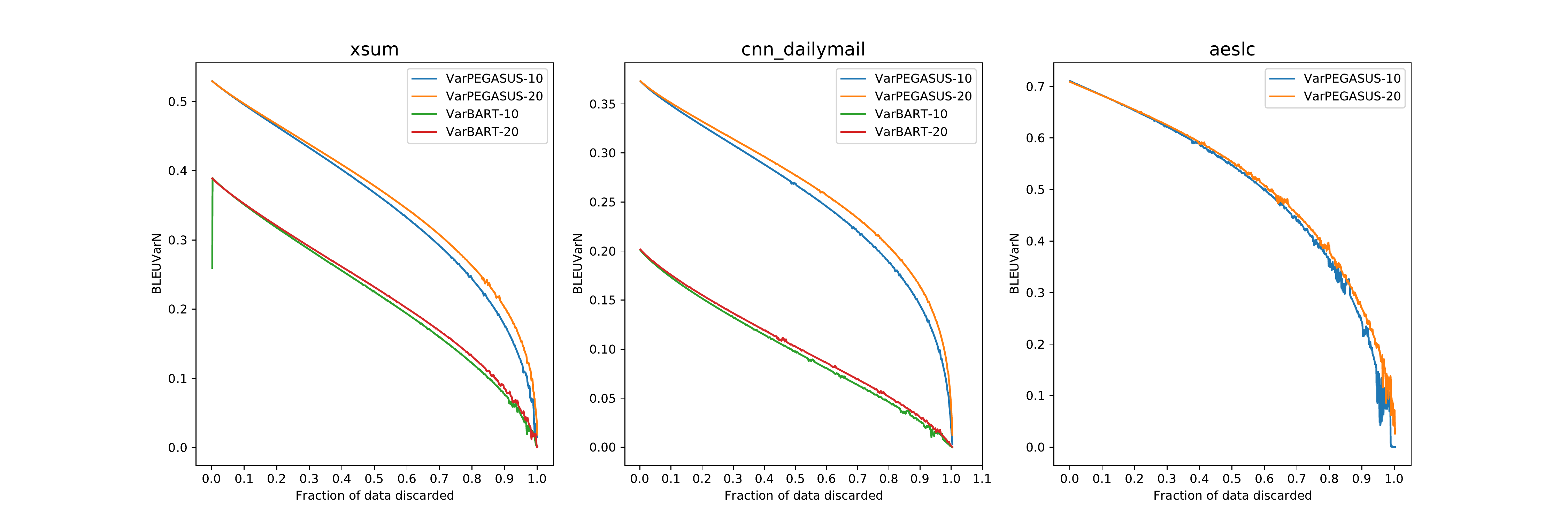}}
    \caption{BLEUVarN curves as a function of data discarded due to low ROUGE-1 scores.}
    \label{fig:bleuvars}
\end{figure*}

To further illustrate how BLEUVarN behaves across different parts of the data, Figure \ref{fig:bleuvars} shows the decrease in the average BLEUVarN of all Variational models as we gradually discard samples with low ROUGE-1 scores from each dataset. We observe that for the samples with the highest ROUGE performance BLEUVarN becomes almost zero. This observation further supports our argument that model uncertainty has a significant impact on model performance. 

\subsubsection{MC-10 vs MC-20}
From Figure \ref{fig:r1_data} we can see that MC dropout with 20 samples performs better than 10 samples, resulting in higher performance. In more detail, for highly uncertain data, both MC-10 and MC-20 converge to similar BLEUVarN values (Figure \ref{fig:bleuvars}) as well as ROUGE scores (Figure  \ref{fig:r1_data}). On the other side of the spectrum, for low uncertainty data, using $20$ samples leads to bigger performance increase along with a little higher BLEUVarN scores.

Based on these observations, we conclude that MC dropout with 20 samples is generally better in terms of performance. This comes at the cost of increased computational overhead because it requires running twice as many stochastic passes with MC dropout. However, this computation is embarrassingly parallelizable in modern hardware, and can be easily optimized by batching MC dropout generations with different dropout masks for each sample within the batch.

Although we have shown that MC dropout with 20 samples performs better than 10 samples, we observed that further increasing this number, for example to 30 or 50 samples, was beginning to bring diminishing returns. Furthermore, the performance increase we got by using 10 and 20 samples was substantial enough while the runtimes for MC dropout with more samples were becoming a lot longer. For these reasons we refrained from increasing it even further in order to keep computational capacity manageable. 

\subsubsection{VarBART vs VarPEGASUS}
Out of the two models, VarPEGASUS is consistently showing the biggest increase in performance as more uncertain samples are dropped from the dataset. It should be noted here, that the decline in performance as data uncertainty increases, is much steeper for VarBART than it is for VarPEGASUS on both the XSum and the CNN/DM dataset. This coincides with the fact that VarPEGASUS also has much higher BLEUVarN uncertainty as shown in Figure \ref{fig:bleuvars}, which hints us that the PEGASUS model is in general less confident about the outputs it generates. Anecdotally, we can say here that PEGASUS is more aware of the things it does not know, and as a result it seems to benefit more from the uncertainty estimates.

\subsection{Bayesian vs deterministic summarization}
The next experiment focuses on the Bayesian summarization method proposed in Section \ref{subsec:bayes_inference}. We compare the performance of Bayesian summarization using the VarBART and VarPEGASUS models against the standard summarization paradigm using the deterministic BART and PEGASUS models. Our goal is to verify the efficacy of Bayesian summarization as a post-hoc way of improving summarization performance.

\begin{table*}[t!]
\caption{\label{tab:varpegasus_baselines} A comparison of our VarBART and VarPEGASUS models against the deterministic BART and PEGASUS.}
\begin{center}
\begin{tabular}{rccccccccc}
\hline
& \multicolumn{3}{c}{\textbf{XSum}} & \multicolumn{3}{c}{\textbf{CNN/DM}} & \multicolumn{3}{c}{\textbf{AESLC}}\\\
{\bf Model} & {\bf R-1} & {\bf R-2} & {\bf R-L} & {\bf R-1} & {\bf R-2} & {\bf R-L} & {\bf R-1} & {\bf R-2} & {\bf R-L}\\\hline
BART & 42.69 & 20.66 & 35.29 & 42.32 & 20.28 & 36.21 & - & - & - \\
VarBART-10 & 42.97 & 20.86 & 35.56 & 42.65 & 20.64 & 36.56 & - & - & - \\
VarBART-20 & \textbf{43.07} & \textbf{20.97} & \textbf{35.68} & \textbf{42.76} & \textbf{20.76} & \textbf{36.69} & - & - & - \\
\hline
PEGASUS & 44.90 & 23.33 & 37.74 & 41.68 & 20.24 & 36.17 & 35.97 & 20.28 & 35.09\\
VarPEGASUS-10 & 44.93 & 23.54 & 38.01 & 42.04 & 20.75 & 36.76 & 36.36 & \textbf{21.40} & \textbf{35.58}\\
VarPEGASUS-20 & \textbf{45.32} & \textbf{23.87} & \textbf{38.29} & \textbf{42.25} & \textbf{20.98} & \textbf{36.94} & \textbf{36.41} & 21.00 & 35.53\\
\hline
\end{tabular}
\end{center}
\end{table*}

Table \ref{tab:varpegasus_baselines}  reports the ROUGE-1, ROUGE-2 and ROUGE-L F-scores of our VarBART and VarPEGASUS models along with the deterministic BART and PEGASUS models on all benchmark datasets, re-evaluated for consistency. The results show that Bayesian summarization is effective, with both VarBART and VarPEGASUS outperforming their deterministic counterparts on all datasets. Furthermore, increasing the number, $N$, of samples generated during the Bayesian inference, improves performance for all datasets except for AESLC, at the cost of increased computational complexity as discussed in Section \ref{subsec:bleuvar_exp}.

Note that our goal in this work was not to compete with other state-of-the-art models. What we want to show is that relying on the agreement between multiple Bayesian summaries for the same input, is an effective way to boost the summarization performance of deterministic models. Also, this is a post-hoc method and does not involve training new models, which makes it easily applicable to many different scenarios.

\begin{figure*}
    \makebox[\textwidth][c]{\includegraphics[width=1.2\textwidth]{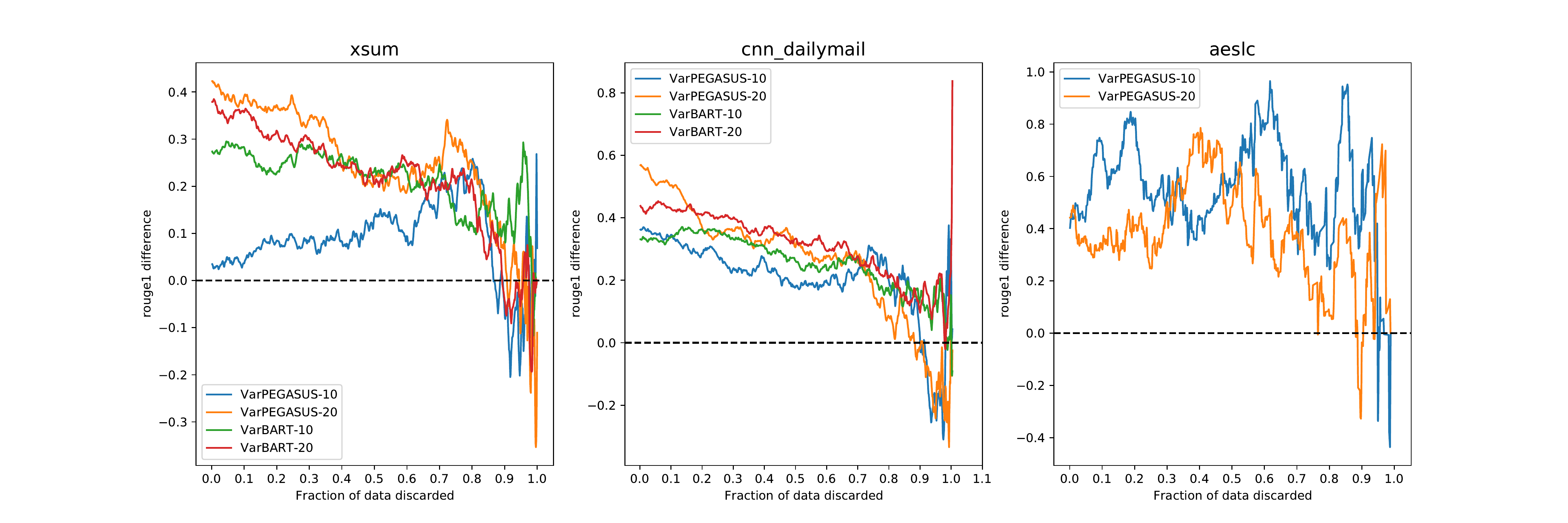}}
    \caption{Difference in ROUGE-1 between Variational models and their deterministic counterparts versus the fraction of data discarded. Positive values indicate that deterministic ROUGE-1 is lower than Variational.}
    \label{fig:r1_diff}
\end{figure*}

Figure \ref{fig:r1_diff} plots the difference in ROUGE-1 of each Variational model with its deterministic counterpart versus the fraction of the data discarded due to high uncertainty. Similar plots for ROUGE-2 and ROUGE-L can be found in Appendix \ref{sec:appendix}. Positive values indicate that the Variational model achieves a higher score than the deterministic one. These plots give us a better view of how the Variational models fare against the deterministic ones for different levels of uncertainty. As far as we know, this is the first study to directly compare Variational and deterministic models on data with varying levels of uncertainty.

Looking at the curves, we clearly see that the differences are positive for most uncertainty levels but start decreasing as more data with high uncertainty are discarded. For the top $10\%-20\%$ most certain samples we start seeing a fluctuation between positive and negative values. This pattern is in line with the observations made in Figure \ref{fig:r1_data}, and leads us to believe that there is a significant gap between the deterministic model's performance on the $20\%$ most certain samples and the rest of the data. 


These observations lead us to the following conclusions. For samples of very low uncertainty, we can expect both Variational and deterministic models to converge to equally good outputs. In contrast, as uncertainty becomes higher, we observe a clear advantage of the Variational summaries over the deterministic ones. This pattern is consistent across all models and datasets, and underpins our case that Bayesian summarization is beneficial for the majority of inputs.

\begin{table}[t!]
\caption{\label{tab:bad_sample} High uncertainty example from the XSum dataset. Sample summary (1), in bold typeface, is the {\em median summary} according to our approach.}
\begin{footnotesize}
\begin{tabular}{p{0.95\linewidth}}
\hline
\textbf{Bayesian samples:}
\begin{enumerate}
    \item {\bf When John Choe launched his first hotel in Singapore, he had no idea what he was getting himself into. (R1: 22.22)}
    \item When Singapore's Frasers Centre hired him as its first managing director, he was told it would take him five years to get off the ground.
    \item In his early 20s, when he was working as a waiter in a luxury hotel in Hong Kong, David Choe always dreamed of running his own business.
    \item "When I was a teenager, I used to say to myself 'I want to start my own company'."
    \item When John Choe was appointed chief executive of a Singapore-based property firm in the early 1990s, he said he wanted to "make a difference to people's lives".
    \item When David Choe was asked if he would ever run a hotel company, he thought it would be a good idea.
    \item "I always wanted to be a hotelier," says Fraser Choe.
    \item As a young entrepreneur with no experience in hospitality, John Choe had no idea what he was about to achieve.
    \item ). 
    \item "When I started the company, I said 'let's see what we can do, let's see what we can achieve, let's see what we can achieve'."
\end{enumerate} \\

\hline
\textbf{Deterministic summary}: When Choe Swee Swee was appointed chief executive of one of Singapore\'s biggest property firms, he told the BBC he wanted to "make the world a better place". (R1: 21.81) \\
\hline
\textbf{Target summary}: On the first day in his new job, Choe Peng Sum was given a fairly simple brief: "Just go make us a lot of money." \\
\hline
\textbf{BLEU variance}: 0.96 \\
\hline
\end{tabular}
\end{footnotesize}
\end{table}

\begin{table}[t!]
\caption{\label{tab:good_sample} Low uncertainty example from XSum. Sample summary (7), in bold typeface, is the {\em median summary} selected according to our approach. In the parentheses we show the ROUGE-1 score for the median Bayesian summary and the deterministic summary.}
\begin{footnotesize}
\begin{tabular}{p{0.95\linewidth}}
\hline
\textbf{Bayesian samples}:
\begin{enumerate}
    \item Torquay United have signed Torquay United have signed Myles Keating.
    \item Torquay United have signed defenders Myles Anderson and Ruairi Keating.
    \item National League side Torquay United have signed defender Lewis Anderson and striker Ruairi Keating.
    \item Torquay United have signed defender Liam Anderson on a deal until the end of the season, while winger Ruairi Keating has joined until the end of the season.
    \item Torquay United have signed defender Matt Anderson on a two-and-a-half-year deal and brought in Republic of Ireland striker Myles Keating on a short-term deal.
    \item Torquay United have signed defender James Anderson and striker Myles Keating.
    \item {\bf Torquay United have signed defender Myles Anderson and striker Ruairi Keating. (R1: 62.5)}
    \item Torquay United have signed defender Lewis Anderson and striker Ruairi Keating.
    \item Torquay United have loaned defender Myles Anderson.
    \item National League strugglers Torquay United have signed defender Lewis Anderson on a two-and-a-half-year deal and Irish striker Ruairi Keating until the end of the season.
\end{enumerate} \\
\hline
\textbf{Deterministic summary}: Torquay United have signed defender Myles Anderson and striker Ruairi Keating until the end of the season. (R1: 52.63) \\
\hline
\textbf{Target summary}: Torquay United have signed Barrow defender Myles Anderson on a permanent deal, and Irish forward Ruairi Keating on non-contract terms. \\
\hline
\textbf{BLEU variance}: 0.38 \\
\hline
\end{tabular}
\end{footnotesize}
\end{table}

\subsection{Qualitative analysis}
In order to better illustrate our findings in this work, we present a couple of real examples from VarPEGASUS-10 on XSum. For each example, we show the 10 sample summaries generated with MC dropout for the same input, as well as the corresponding BLEUVarN score. We have highlighted the median summary in bold typeface and for the sake of comparison we also show the summary generated by the deterministic PEGASUS model.

The first example (Table \ref{tab:bad_sample}) is a case of high uncertainty from the XSum dataset. We can see that all 10 samples are considerably different from one another, which leads to a high BLEUVarN score. In contrast, the second example (Table \ref{tab:good_sample}) has much lower uncertainty. In this case all 10 samples seem to mostly agree on the main points and as a result BLEUVarN is fairly low. Here, the median summary is the one that represents better this agreement.

Looking at the ROUGE-1 score for both examples we can see there's a rather drastic difference as well. For the sample in Table \ref{tab:bad_sample} we can see that neither the deterministic nor the Bayesian summary show a strong performance, yet even in that case the median Bayesian summary scores a bit higher. On the other hand, the sample in Table \ref{tab:good_sample} showcases a solid performance from both the deterministic and the Bayesian summary. Here it is evident that the median Bayesian summary is close but slightly better than the deterministic summary in terms of ROUGE.


\section{Conclusions and future work}
\label{sec:conclusion}
This work explored Bayesian methods in the context of text summarization. We extended state-of-the-art summarization models with MC dropout to approximate Bayesian inference, and demonstrated how BLEUVarN can be used to quantify model uncertainty. This allows us to effectively identify high uncertainty summaries at prediction time, which can be a significant advantage.

Furthermore, we show that ensembling multiple stochastic summaries generated by Variational Bayesian models can lead to improved performance compared to similar deterministic models. This finding is verified by experiments for two different models and across 3 benchmark datasets.

It should be noted here that the proposed methods are directly applicable to other abstractive summarization datasets as well. We acknowledge that some of the more interesting summarization problems involve much longer summaries, for example scientific abstracts. In this work we focused on datasets of short summaries in order to be more resource efficient and conduct more experiments. There's a lot of interesting work that focuses on the summarization of longer documents \cite{Gidiotis2020ADocuments, Zaheer2020BigSequences} that could potentially be applied in combination with the methods we propose here.

Our work can have a broader impact in several ways. To the research community, being the first work to study Bayesian uncertainty for abstractive summarization and paving the way for other similar methods. To the industry, because it improves automatic summarization systems and can be paired nicely with active learning and human-in-the-loop approaches. Finally, to the end users, improving their experience and building up confidence towards automatic summarization systems.

\bibliographystyle{unsrtnat}
\bibliography{bayesian_summarization}  






\appendix

\section{Appendix}
\label{sec:appendix}

\begin{figure*}
    \makebox[\textwidth][c]{\includegraphics[width=1.2\textwidth]{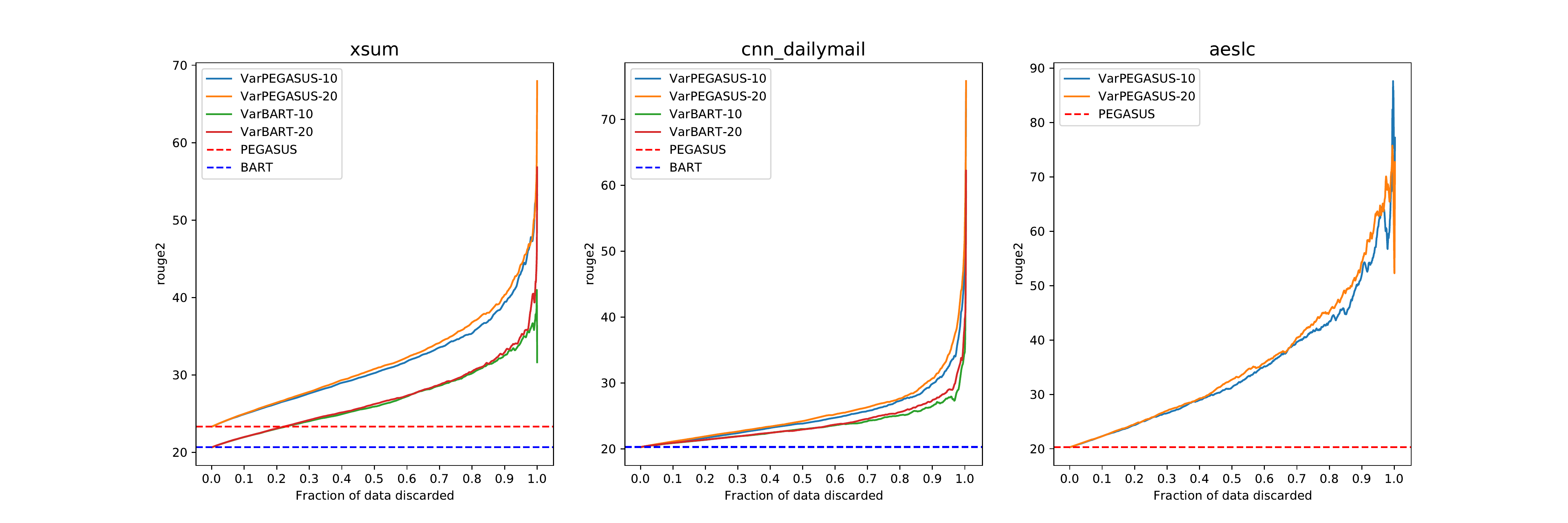}}
    \caption{ROUGE-2 scores vs fraction of data discarded due to high BLEUVarN. The straight dashed lines indicate the performance level of the deterministic PEGASUS and BART models.}
    \label{fig:r2_data}
\end{figure*}

\begin{figure*}
    \makebox[\textwidth][c]{\includegraphics[width=1.2\textwidth]{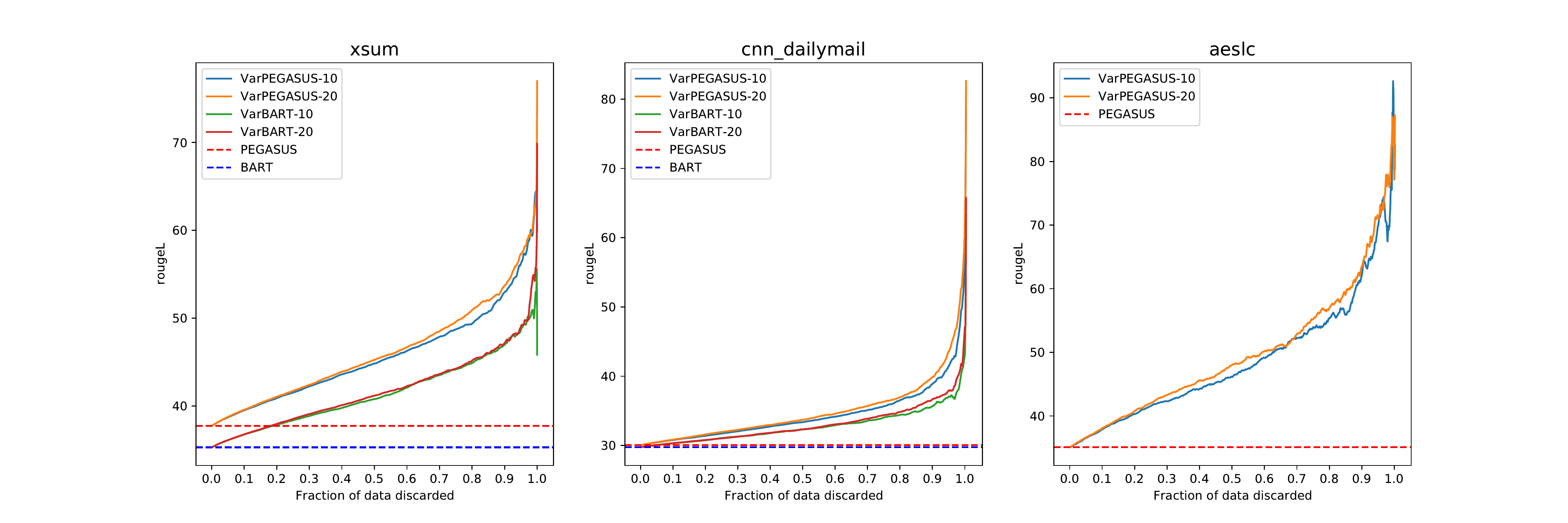}}
    \caption{ROUGE-L scores vs fraction of data discarded due to high BLEUVarN. The straight dashed lines indicate the performance level of the deterministic PEGASUS and BART models.}
    \label{fig:rl_data}
\end{figure*}

Figures \ref{fig:r2_data} and \ref{fig:rl_data} show the performance versus data retention curves of our Variational models in terms of ROUGE-2 and ROUGE-L F-score respectively. The observations here are similar to Figure \ref{fig:r1_data}.

\begin{figure*}
    \makebox[\textwidth][c]{\includegraphics[width=1.2\textwidth]{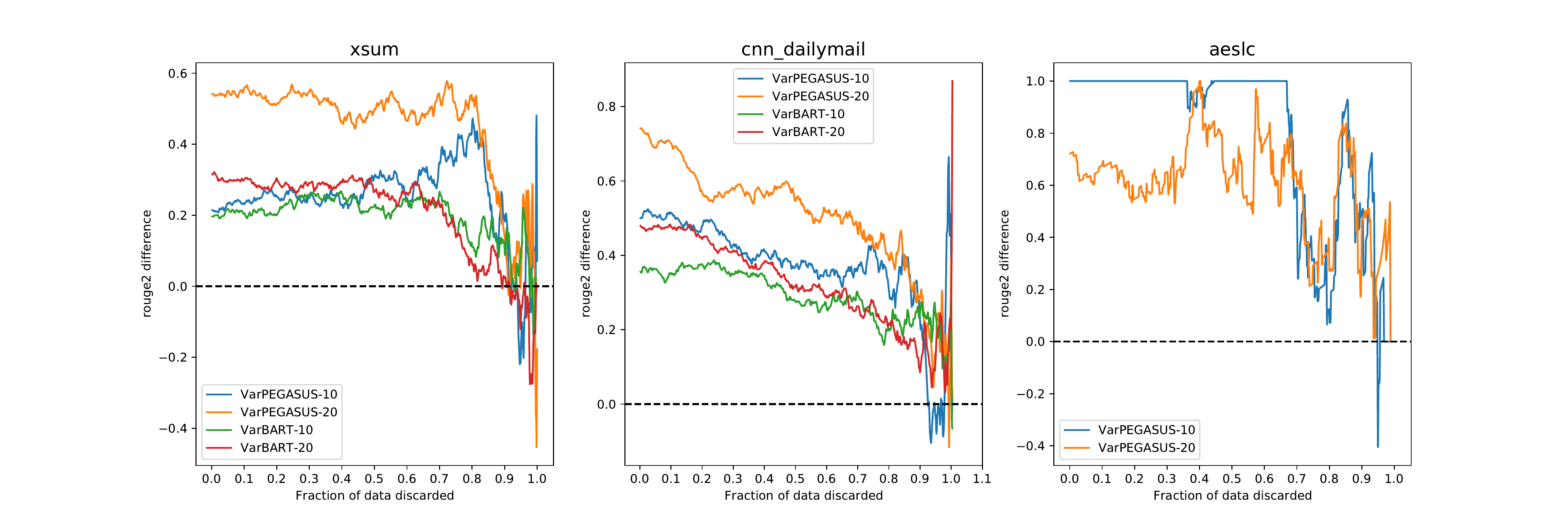}}
    \caption{Difference in ROUGE-2 between Variational models and their deterministic counterparts versus the fraction of data discarded. Positive values indicate that deterministic ROUGE-2 is lower than Variational.}
    \label{fig:r2_diff}
\end{figure*}

\begin{figure*}[t!]
    \makebox[\textwidth][c]{\includegraphics[width=1.2\textwidth]{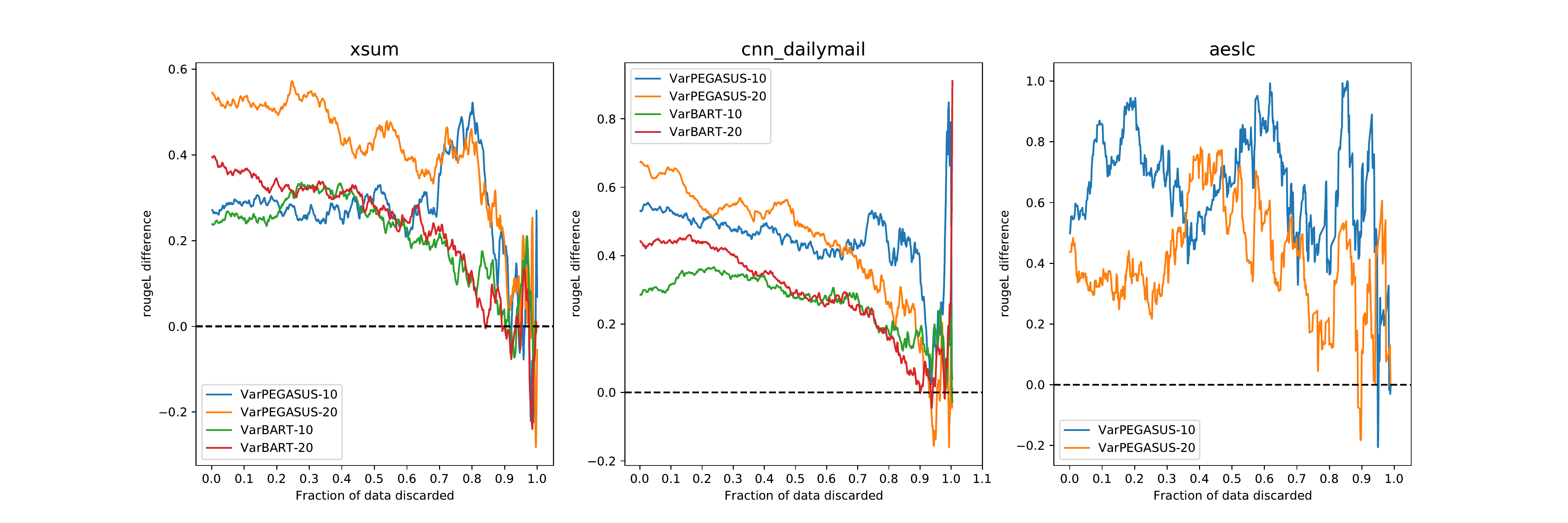}}
    \caption{Difference in ROUGE-L between Variational models and their deterministic counterparts versus the fraction of data discarded. Positive values indicate that deterministic ROUGE-L is lower than Variational.}
    \label{fig:rl_diff}
\end{figure*}

Figures \ref{fig:r2_diff} and \ref{fig:rl_diff} show the differences in ROUGE-2 and ROUGE-L performance of the Variational models versus the deterministic ones. What we see here is in aggreement with Figure \ref{fig:r1_diff}.

\end{document}